\documentclass[letterpaper]{article} 
\usepackage{aaai2026}  
\usepackage{times}  
\usepackage{helvet}  
\usepackage{courier}  
\usepackage[hyphens]{url}  
\usepackage{graphicx} 
\usepackage{natbib}  
\usepackage{caption} 
\usepackage{algorithm}
\usepackage{algorithmic}
\usepackage{amsmath}
\usepackage{multirow}
\usepackage{bm}
\usepackage{amsfonts}
\usepackage{graphicx}
\usepackage{caption}
\usepackage{booktabs}
\usepackage{subcaption}
\usepackage{newfloat}
\usepackage{listings}
\DeclareCaptionStyle{ruled}{labelfont=normalfont,labelsep=colon,strut=off} 
\floatstyle{ruled}
\newfloat{listing}{tb}{lst}{}
\floatname{listing}{Listing}
\title{AAAI Press Formatting Instructions \\for Authors Using \LaTeX{} --- A Guide}
\author {
    Yuting Tang\textsuperscript{\rm 1}, 
    Weibang Jiang\textsuperscript{\rm 1,\rm 2}, 
    Shanglin Li\textsuperscript{\rm 3}, 
    Yong Li\textsuperscript{\rm 4}, 
    Chenyu Liu\textsuperscript{\rm 1}, 
    Xinliang Zhou\textsuperscript{\rm 1},
    Yi Ding\textsuperscript{\rm 1}\thanks{Corresponding authors}, 
    Cuntai Guan\textsuperscript{\rm 1}\footnotemark[1]
}
\affiliations {
    \textsuperscript{\rm 1}College of Computing and Data Science, Nanyang Technological University\\
    \textsuperscript{\rm 2}Shanghai Jiao Tong University\\
    \textsuperscript{\rm 3}Advanced Telecommunications Research Institute International \\
    \textsuperscript{\rm 4}Southeast University \\ 
    yuting.tang@ntu.edu.sg, 935963004@sjtu.edu.cn, shanglin@atr.jp, mysee1989@gmail.com, \{chenyu003, xinliang001\}@e.ntu.edu.sg, \{ding.yi, ctguan\}@ntu.edu.sg
}

\usepackage{bibentry}

\title{EEG-DLite: Dataset Distillation for Efficient Large EEG Model Training}

\begin{document}
\maketitle

\begin{abstract}
Large-scale EEG foundation models have shown strong generalization across a range of downstream tasks, but their training remains resource-intensive due to the volume and variable quality of EEG data. In this work, we introduce EEG-DLite, a data distillation framework that enables more efficient pre-training by selectively removing noisy and redundant samples from large EEG datasets. EEG-DLite begins by encoding EEG segments into compact latent representations using a self-supervised autoencoder, allowing sample selection to be performed efficiently and with reduced sensitivity to noise. Based on these representations, EEG-DLite filters out outliers and minimizes redundancy, resulting in a smaller yet informative subset that retains the diversity essential for effective foundation model training. Through extensive experiments, we demonstrate that training on only 5 percent of a 2,500-hour dataset curated with EEG-DLite yields performance comparable to, and in some cases better than, training on the full dataset across multiple downstream tasks. To our knowledge, this is the first systematic study of pre-training data distillation in the context of EEG foundation models. EEG-DLite provides a scalable and practical path toward more effective and efficient physiological foundation modeling.


\end{abstract}

\section{Introduction}

Brain-computer interfaces (BCIs) enable direct communication between the human brain and external devices, unlocking a wide range of applications in neurorehabilitation~\cite{mtp-bci}, assistive technologies~\cite{Tang2024}, and cognitive monitoring~\cite{zhou2025brainfoundationmodelssurvey}. Among the various modalities used in BCIs, electroencephalography (EEG) is the most widely adopted due to its non-invasiveness and high temporal resolution. However, EEG signals are inherently noisy, high-dimensional, and subject to significant inter-subject variability, which presents challenges for learning robust and generalizable representations~\cite{WangNth2025,carzaniga2025the}.

To address these challenges, recent advances have introduced EEG foundation models, which are large-scale neural networks pre-trained on extensive unlabeled EEG datasets using self-supervised objectives \cite{zhou2025brainfoundationmodelssurvey}. These models, often built on convolutional backbones or Transformer-based architectures, can exceed 400 million parameters. Typically, they are first pre-trained on large-scale EEG data from diverse BCI tasks and later fine-tuned for specific downstream applications. 
These models demonstrate competitive performance across various downstream tasks while maintaining a unified architectural framework \cite{jiang2024largebrainmodellearning, eegpt, wang2025cbramod}.
While effective, this pre-training paradigm is computationally expensive, often requiring significant GPU time, storage, and energy~\cite{jiang2024largebrainmodellearning}. The high cost also limits the feasibility of further exploration into parameter optimization and neural architecture search~\cite{yu2023datasetdistillationcomprehensivereview}.

\begin{figure}[!t]
    \centering
    \includegraphics[width=\linewidth]{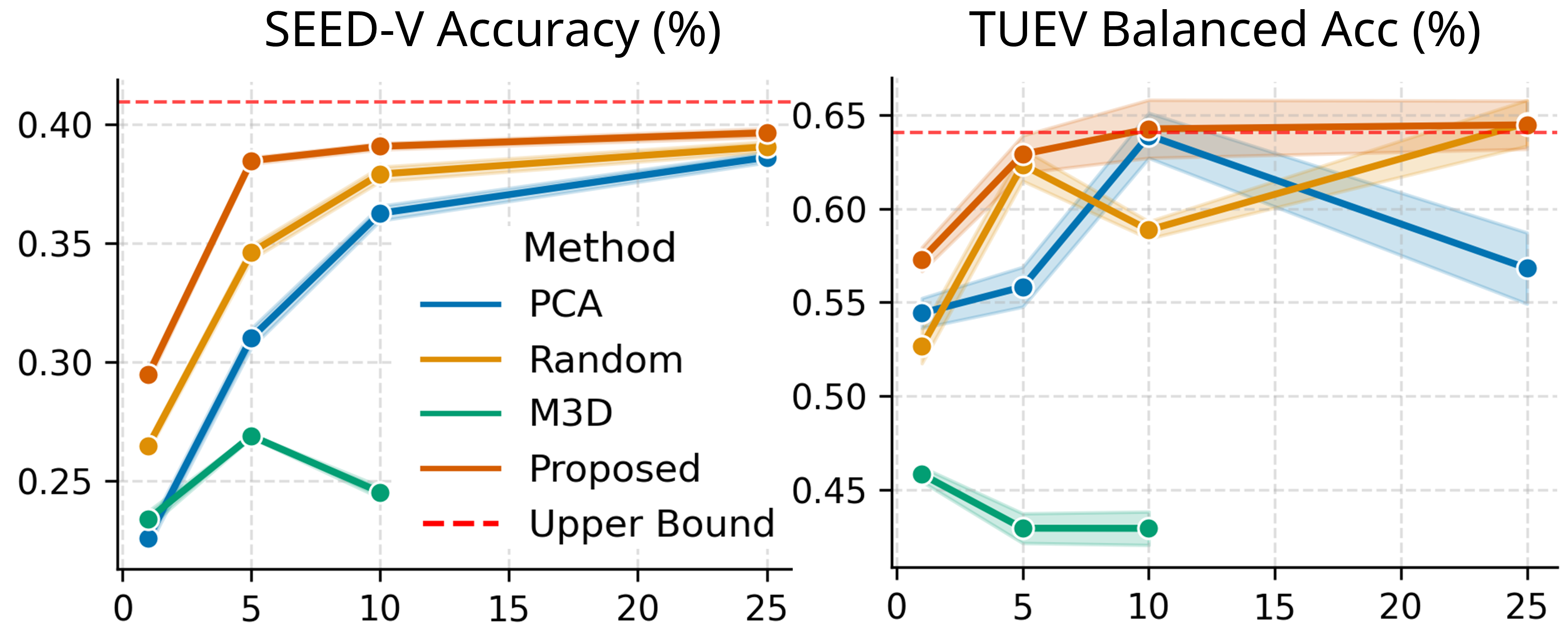}
\caption{Performance trends of LaBraM on downstream tasks using pre-training datasets distilled by EEG-DLite at different ratios $\eta$.}
    \label{fig:first-figure}
\end{figure}

Recent advances in computer vision have shown that it is possible to significantly reduce the computational burden of large-scale model training by distilling extensive datasets into compact, information-rich subsets~\cite{zhang2024m3ddatasetcondensationminimizing,joshi2024datasetdistillationknowledgedistillation,killamsetty2021gradmatchgradientmatchingbased}. Motivated by the broader goal of data-efficient pre-training, we turn our focus to EEG data, which presents unique and domain-specific challenges that demand tailored solutions. First, EEG signals exhibit a low signal-to-noise ratio (SNR), as neural activity is easily obscured by artifacts from eye movements, muscle activity, and external interference~\cite{9441341}. Even after preprocessing, residual noise can persist and adversely impact model learning. Second, EEG recordings often contain significant redundancy; temporally adjacent segments may capture overlapping or repetitive patterns, offering limited new information~\cite{amin2016novel}. These characteristics make EEG fundamentally different from vision data in structure and signal quality, necessitating data-efficient strategies that directly address its inherent noise and redundancy. Despite these challenges, little is known about how the volume and composition of pre-training EEG data influence the generalization ability of foundation models, leaving a critical gap in the development of scalable and effective large-scale EEG modeling.

To overcome these limitations, we propose \textbf{EEG-DLite}, a \textit{data distillation framework} for data-efficient pre-training of large EEG foundation models. EEG-DLite systematically prunes large, unlabeled EEG datasets that are collected using varied channel montages into smaller, more representative subsets. Due to the high dimensionality and low signal-to-noise ratio of EEG signals, instead of selecting samples directly using EEG signals, our approach begins by encoding EEG segments into compact latent representations using a self-supervised autoencoder. Based on these representations, we apply two key techniques:
\begin{itemize}
    \item A robust \textit{outlier filtering mechanism} to remove noisy or corrupted EEG segments
    \item A \textit{divergence-based redundancy reduction} method to reduce data redundancy while preserving informative diversity
\end{itemize}

These steps produce a significantly smaller dataset that still captures the diversity and structure necessary for effective model training. In our experiments, training on only \textbf{5\% of a 2,500-hour EEG dataset} distilled using EEG-DLite achieves \textit{comparable}, or \textit{even superior}, performance on several downstream tasks compared to training on the full dataset, as shown in Figure~\ref{fig:first-figure}. Additionally, this reduces GPU pre-training time from \textbf{30 hours to just 2 hours}, under the same hardware conditions.

To our knowledge, this is the \textbf{first study to explore data distillation for physiological signals} like EEG in the context of large foundation models. We also perform systematic comparisons of \textit{generative} and \textit{selection-based} distillation approaches, providing new insights into what makes pre-training data effective in EEG modeling\footnote{The code is available at https://github.com/t170815518/EEG-DLite}. 

The main contributions of this work are as follows:
\begin{itemize}
    \item We propose \textit{the first data distillation framework tailored for large-scale EEG foundation model pre-training}, achieving comparable or even superior performance using only 5\% of the original training data.
    
    \item We present the first comparison between synthetic data generation and selection-based distillation on EEG. The results show the challenges EEG poses on the generation approach and highlight the effectiveness of a core-set selection approach.
    
    \item We systematically analyze how the quantity of pre-training data affects model generalization, providing empirical evidence for the efficiency and robustness of our distilled EEG subsets.
\end{itemize}

\begin{algorithm}[!t]
\caption{{Overview of the EEG-DLite Framework}}
\begin{algorithmic}[1]
\label{alg:overview}
\REQUIRE EEG dataset $\mathcal{X} = \{X_i \in \mathbb{R}^{C \times T}\}_{i=1}^N$, encoder $\xi$, decoder $\mathcal{D}$, latent representation dimension $d$, outlier threshold $\tau$, distillation ratio $\eta$
\ENSURE Distilled EEG subset $\mathcal{S} \subset \mathcal{X}$

\STATE Compute spectral views $X_i^{(m)}, X_i^{(\phi)} \leftarrow \texttt{FFT}(X_i)$ and concatenate: $X_i^{\text{all}} = [X_i, X_i^{(m)}, X_i^{(\phi)}]$
\STATE Train $(\xi, \mathcal{D})$ to minimize: 
\[
\mathcal{L} = \mathcal{L}_{\text{Rec}}+ \beta \mathcal{L}_{\text{IDC}}
\]
\STATE Encode each $X_i^{\text{all}}$ to $z_i = \xi(X_i^{\text{all}})$, collect $\mathcal{Z} = \{z_i\}_{i=1}^N$
\STATE Compute outlier scores: $\sum_{i=1}^d \log\frac{1}{p_i(x_i) + \alpha}$

\STATE Remove top $\tau\%$ OODs: $\mathcal{Z}' \subset \mathcal{Z}$, $\mathcal{X}' \subset \mathcal{X}$
\STATE Select $\eta\%$ with diversity sampling:
\[
    \min_{\bm\mu \subset \mathcal{Z'}}\max_{\mathbf{z}\in \mathcal{Z'}}\min_{k\in \mathcal{K}}  ||\mathbf{z} - \bm{\mu}_k||_2^2.
\]
\STATE Return subset $\mathcal{S} \subset \mathcal{X'}$
\end{algorithmic}
\end{algorithm}

\section{Related Work}
\begin{figure*}[t]
    \centering
    \includegraphics[width=0.95\linewidth]{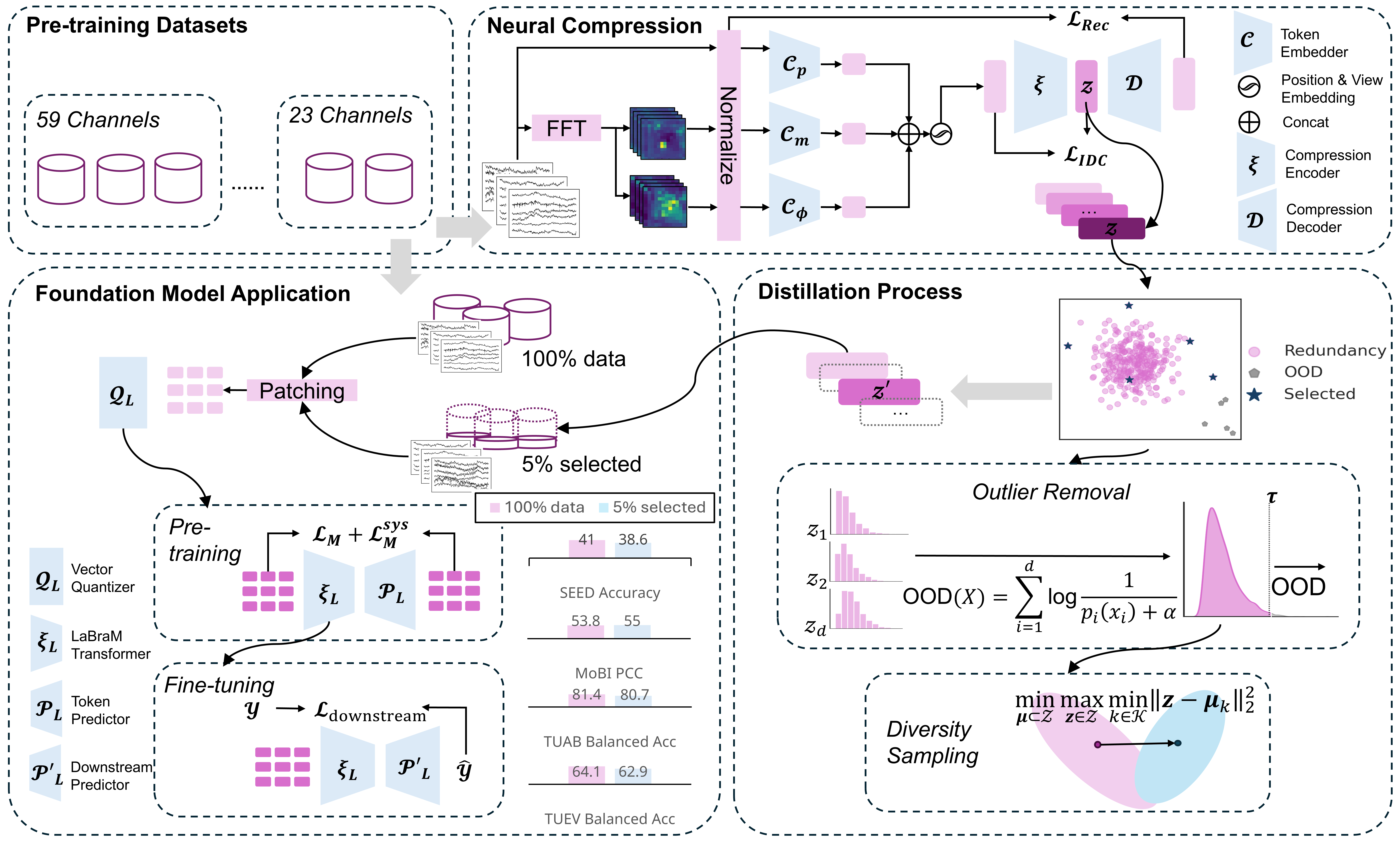}
\caption{
Overview of the proposed framework. In the first stage, an auto-encoder is trained with self-supervised learning to project each EEG segment $X$ into a lower-dimensional latent representation $\mathbf z$. In the second stage, out-of-distribution scores are computed for all samples, and the top $\tau\%$ with the highest scores are excluded. Finally, $\eta\%$ of the datasets is selected via diversity sampling to retain only the most informative data. }
    \label{fig:framework_overview}
\end{figure*}

\subsection{EEG foundation models} 
EEG foundation models have been an emerging paradigm for decoding and modeling brain activity using deep neural networks pretrained on large-scale EEG recordings~\cite{zhou2025brainfoundationmodelssurvey,lai2025simplerevieweegfoundation}. Typical architectures adopt self-supervised learning (SSL) objectives and employ CNNs, transformers, or hybrid architectures to extract generalizable neural representations. These models have shown strong performance on downstream tasks such as cognitive state decoding and clinical classification~\cite{jiang2024largebrainmodellearning, eegpt}. However, training them requires massive datasets, often exceeding millions of EEG segments, which poses significant computational and storage burdens. More critically, existing efforts largely overlook the influence of training data composition and selection on model performance. While prior work has focused on architectural innovation and transfer learning strategies, few have explored how data quality, representativeness, and redundancy affect the foundation model's generalization ability. Our work addresses this gap by introducing a data-centric framework that selects diverse, informative EEG segments to reduce training costs without compromising generalization. By emphasizing principled data selection, we aim to enhance both the scalability and explainability of large EEG models, bridging the methodological gap between neuroscience and AI through efficient and standardized data curation.

\subsection{Data distillation}
Data distillation refers to the process of constructing a compact and informative subset of training data that allows a model to achieve comparable performance to that trained with the full dataset, while significantly reducing computational costs. Existing distillation methods generally fall into two main categories: synthetic data generation and core-set selection~\cite{joshi2024datasetdistillationknowledgedistillation, sener2018active}. Synthetic data generation focuses on generating samples that simulate the learning behavior or statistical properties of the original dataset. They have gained popularity in many computer vision tasks due to the flexibility and expressive capacity~\cite{ding2024time}. These methods often use optimization objectives such as performance matching, parameter matching, or distribution matching~\cite{yu2023datasetdistillationcomprehensivereview, zhang2024m3ddatasetcondensationminimizing}. However, they typically rely on bi-level optimization, which involves nested optimization loops, making them computationally intensive. Core-set selection, in contrast, selects a real and representative subset from the training data~\cite{sener2018active}. They are more stable and interpretable since they work with real data, but their selection process becomes increasingly expensive as both data size and dimensionality grow. This is particularly problematic for EEG signals, which are high-dimensional, noisy, and vary across subjects. To address the computational burden, we adopt SSL to train a lightweight autoencoder that compresses EEG segments into low-dimensional latent representations, enabling efficient selection \cite{balestriero2023cookbookselfsupervisedlearning}.

\section{Methodology}
This section elaborates on the details of the proposed framework, EEG-DLite, which is a model-agnostic EEG data distillation framework. It is designed to select a compact yet representative subset from massive unlabeled data for efficient pre-training. As illustrated in Figure~\ref{fig:framework_overview} and Algorithm~\ref{alg:overview}, the framework consists of three main steps: (1) a self-supervised multi-view autoencoder compresses high-dimensional EEG signals into low-dimensional latent representations; (2) an outlier detection module filters out anomalous or low-quality samples that may degrade diversity; and (3) a $k$-center selection algorithm identifies a diverse set of EEG segments that best represent the original dataset.

\subsection{Multi-view Neural Compressor}
To handle the high dimensionality of EEG signals, we train a lightweight autoencoder with SSL that maps raw EEG segments into compact latent representations without requiring labeled data \cite{VAEEG, zhang2023selfsupervisedtimeseriesrepresentation}. This architecture is intentionally decoupled from any specific foundation model to ensure flexibility and broad compatibility.

\paragraph{Model Architecture}
Each EEG segment $X \in \mathbb{R}^{C \times T}$, where $C$ denotes the number of channels and $T$ denotes the temporal length, is processed together with its spectral counterpart obtained via Fast Fourier Transform (FFT). Both the raw and spectral views are partitioned into non-overlapping temporal patches of length $T_W$, denoted as $\bm{x}_i \in \mathbb{R}^{C \times T_W}$. These patches in the potential, magnitude, and phase domains are individually encoded by convolutional neural networks $\mathcal{C}_p$, $\mathcal{C}_m$, and $\mathcal{C}_\phi$ to extract localized spatiotemporal features and produce token embeddings. The resulting token sequences are enriched with positional and view-specific embeddings and passed to a transformer-based encoder, denoted by $\xi$, which captures global dependencies across the sequence. A segment-level representation $\mathbf z$ is then computed by averaging all token embeddings. Finally, a decoder $\mathcal{D}$, which consists of a shallow transformer block followed by multi-layer perceptron layers, reconstructs both the original signals and their spectral components from the compressed token representations.

\paragraph{Optimization Objective} The SSL objective includes two components: a reconstruction loss $\mathcal{L}_\text{Rec}$ and an inter-instance discrimination loss $\mathcal{L}_{\text{IDC}}$. The reconstruction loss minimizes the mean squared error between input patches $\bm{x}_i$ with length $L$ and reconstructions $\bm{x}_i^\prime$, ensuring the encoder can accurately encode neural signal content. The loss of each sample $X$ is defined as

\begin{equation}
\label{eqn:rec}
    \mathcal{L}_\text{Rec} = \sum_{i=1}^L \left( \bm{x}_i' - \bm{x}_i\right)^2.
\end{equation}

$\mathcal{L}_{\text{IDC}}$ penalizes excessive token similarity between samples in one batch, encouraging feature diversity~\cite{chen2021iceinterinstancecontrastiveencoding}. Specifically, denoting $\mathbf{z}_i$ and ${\mathbf{z}_i'}$ the $i$-th token before and after the encoder, the network projects them into the same embedding space with two separate projectors $g_1$ and $g_2$ and penalizes the cosine similarity between an original token and the encoded tokens in other samples, as described below,
$$
    \mathcal{L}_\text{IDC}=\frac{\log\sum_{i=1}^{|B|} \sum_{j=1, i\neq j}^{|B|}\exp[\operatorname{sim}(g_1(\mathbf{z}_i), g_2(\mathbf{z}_j'))]}{|B|\times(|B|-1)}, 
$$ where $B$ represents a sample batch, $\operatorname{sim}(\cdot)$ represents the cosine similarity. Therefore, the final optimization objective is defined as 
\begin{equation}
\mathcal{L} = \mathcal{L}_{\text{Rec}} + \beta \cdot \mathcal{L}_{\text{IDC}}.
\label{eqn:training-obj}
\end{equation}

\begin{table*}[t]
\centering

\resizebox{\linewidth}{!}{
\begin{tabular}{lcccc|cccc}
\toprule
\multirow{2}{*}{\textbf{Method}} 
& \multirow{2}{*}{$\bm\eta \ (\%)$} 
& \multicolumn{3}{c|}{\textbf{SEED-V}} 
& \multicolumn{3}{c}{\textbf{MoBI}} \\
\cmidrule(lr){3-5} \cmidrule(lr){6-8}
& & Accuracy (\%) & $\bm\kappa$ (\%) & F1 (\%) 
& $PCC$ & $R^2$ & RMSE ($\times 10^2$) \\
\midrule

\multirow{4}{*}{Random} 
& 1  & $26.5\pm0.2$ & $7.4\pm0.3$  & $26.6\pm0.2$ & $0.468\pm0.009$ & $0.180\pm0.007$ & $13.97\pm0.08$ \\
& 5  & $34.6\pm0.3$ & $18.3\pm0.4$ & $34.9\pm0.3$ & $0.530\pm0.015$ & $0.260\pm0.007$ & $13.26\pm0.06$ \\
& 10 & $37.9\pm0.3$ & $22.4\pm0.4$ & $38.3\pm0.3$ & $0.540\pm0.003$ & $0.267\pm0.004$ & $13.22\pm0.03$ \\
& 25 & $39.1\pm0.2$ & $23.9\pm0.3$ & $39.4\pm0.2$ & $0.538\pm0.003$ & $0.263\pm0.005$ & $13.29\pm0.05$ \\

\midrule
\multirow{4}{*}{M3D} 
& 1  & $23.4\pm0.3$ & $3.2\pm0.4$ & $22.7\pm0.4$ & $0.305\pm0.003$ & $0.016\pm0.004$ & $15.06\pm0.04$ \\
& 5  & $26.9\pm0.1$ & $7.9\pm0.1$ & $26.9\pm0.1$ & $0.465\pm0.004$ & $0.171\pm0.004$ & $14.04\pm0.03$ \\
& 10 & $24.5\pm0.2$ & $4.9\pm0.3$ & $24.3\pm0.2$ & $0.500\pm0.009$ & $0.209\pm0.011$ & $13.74\pm0.08$ \\
& 25 &  - & - & - & - & - & - \\

\midrule
\multirow{4}{*}{PCA + DS} 
& 1  & $22.6\pm0.4$ & $2.2\pm0.5$ & $22.2\pm0.4$ & $0.356\pm0.004$ & $0.046\pm0.005$ & $14.90\pm0.03$ \\
& 5  & $31.0\pm0.4$ & $13.4\pm0.5$ & $31.2\pm0.4$ & $0.534\pm0.004$ & $0.262\pm0.000$ & $13.32\pm0.06$ \\
& 10 & $36.3\pm0.3$ & $20.1\pm0.4$ & $36.5\pm0.3$ & $\bm{0.541\pm0.002}$ & $0.265\pm0.003$ & $13.26\pm0.02$ \\
& 25 & $38.6\pm0.2$ & $23.5\pm0.3$ & $39.0\pm0.2$ & $0.546\pm0.001^\dagger$ & $0.276\pm0.004^\dagger$ & $13.15\pm0.03^\dagger$ \\

\midrule
\multirow{4}{*}{Proposed} 
& 1  & $\bm{29.7\pm0.3}^\dagger$ & $\bm{11.3\pm0.3}^\dagger$ & $\bm{29.7\pm0.3}^\dagger$ & $\bm{0.506\pm0.007}^\dagger$ & $\bm{0.225\pm0.010}^\dagger$ & $\bm{13.61\pm0.10}^\dagger$ \\
& 5  & $\bm{38.6\pm0.2}^\dagger$ & $\bm{23.1\pm0.2}^\dagger$ & $\bm{38.9\pm0.2}^\dagger$ & $\bm{0.550\pm0.001}^\dagger$ & $\bm{0.283\pm0.001}^\dagger$ & $\bm{13.15\pm0.07}^\dagger$ \\
& 10 & $\bm{39.1\pm0.1}^\dagger$ & $\bm{23.9\pm0.2}^\dagger$ & $\bm{39.5\pm0.2}^\dagger$ & $\bm{0.541\pm0.002}$ & $\bm{0.277\pm0.014}$ & $\bm{13.07\pm0.04}^\dagger$ \\
& 25 & $\bm{39.7\pm0.2}^\dagger$ & $\bm{24.7\pm0.2}^\dagger$ & $\bm{40.1\pm0.2}^\dagger$ & $\bm{0.550\pm0.003}^\dagger$ & $\bm{0.281\pm0.005}^\dagger$ & $\bm{13.14\pm0.06}^\dagger$ \\

\midrule
\multirow{1}{*}{Full data}
& 100 & $41.0\pm0.6$ & $26.1\pm0.8$ & $41.2\pm0.6$
& $0.538\pm0.010$ & $0.288\pm0.003$ & $12.25\pm0.03$ \\

\bottomrule
\end{tabular}
}

\caption{Performance comparison across different distillation ratios ($\eta$) on SEED-V and MoBI datasets. $\bm\kappa$ refers to Cohen’s kappa coefficient. Bold values indicate the best performance within each distillation ratio, excluding the full data case. DS refers to the diversity sampling. The $^\dagger$ symbol indicates the result is significantly better ($p<0.05$) than the corresponding Random baseline at the same $\eta$ based on Mann–Whitney U test.}
\label{tab:distillation-results}
\end{table*}

\subsection{Outlier Sample Removal}
At this stage, the goal is to remove isolated and unrepresentative samples within the dataset. The operation is conducted in the compressed representation space to ensure efficiency. Specifically, we adopt the Histogram-Based Outlier Score (HBOS) method, which identifies anomalies based on their probabilistic rarity in each feature dimension $d$. 

Each sample $X$ can be assigned with an out-of-distribution (OOD) score, defined as Equation \ref{eqn:hbos}. The top $\tau$ percent of samples are excluded from the subsequent diverse selection step to prevent degrading the diversity and quality of the distilled subset.

\begin{equation}
\label{eqn:hbos}
    \text{OOD}(X) = \sum_{i=1}^d \log\frac{1}{p_i(x_i) + \alpha}
\end{equation}

\subsection{Diversity Sampling}

\begin{table*}[t]
\centering
\resizebox{\linewidth}{!}{
\begin{tabular}{lcccc|cccc}
\toprule
\multirow{2}{*}{\textbf{Method}} 
& \multirow{2}{*}{$\bm\eta \ (\%)$} 
& \multicolumn{3}{c|}{\textbf{TUEV}} 
& \multicolumn{3}{c}{\textbf{TUAB}} \\
\cmidrule(lr){3-5} \cmidrule(lr){6-8}
& & Balanced Acc (\%) & $\bm\kappa$ (\%) & F1 (\%) 
& Balanced Acc (\%) & PR AUC (\%) & AUROC (\%)  \\
\midrule

\multirow{4}{*}{Random} 
& 1& $52.7\pm0.9$ & $46.8\pm0.9$& $73.6\pm0.6$&  $79.5\pm0.1$& $87.7\pm0.1$& $88.2\pm0.1$ \\
& 5&  $62.3\pm0.9$ & $58.1\pm0.9$& $79.3\pm0.6$&  $\bm{80.7\pm0.2}$& $89.9\pm0.2$& $90.0\pm0.1$ \\
& 10&  $58.9\pm0.5$ & $56.7\pm1.5$& $78.9\pm0.7$&  $81.2\pm0.1$& $90.4\pm0.2$& $90.6\pm0.2$ \\
& 25& $64.4\pm1.2$ & $61.5\pm1.5$& $81.2\pm0.7$&  $81.1\pm0.2$& $90.2\pm0.1$& $90.3\pm0.1$ \\

\midrule
\multirow{4}{*}{M3D} 
& 1& $45.8\pm0.4$ & $37.6\pm0.6$ & $68.1\pm0.4$ & $77.3\pm0.1$& $83.4\pm0.2$& $85.2\pm0.1$ \\
& 5& $42.9\pm0.8$ & $42.4\pm0.8$ & $71.2\pm0.4$ & $79.5\pm0.1$& $86.2\pm0.1$& $88.1\pm0.1$ \\
& 10& $42.9\pm0.9$ & $38.9\pm1.1$ & $69.1\pm0.7$ & $80.1\pm0.1$& $86.9\pm0.3$& $88.3\pm0.2$ \\
& 25&  - & - & - & -& - & -\\

\midrule
\multirow{4}{*}{PCA + DS} 
& 1& $54.4\pm0.8^\dagger$ & $48.0\pm0.5^\dagger$& $73.9\pm0.3$& $77.8\pm0.1$& $85.0\pm0.2$& $86.2\pm0.1$ \\
& 5&  $55.8\pm1.1$ & $51.0\pm1.0$& $75.7\pm0.6$&  $81.3\pm0.1^\dagger$& $\bm{89.9\pm0.1}$& $\bm{90.6\pm0.1}^\dagger$ \\
& 10&  $63.9\pm1.2^\dagger$ & $61.6\pm1.9^\dagger$& $81.2\pm1.0\dagger$&  $81.0\pm0.1$& $90.1\pm0.1$& $90.2\pm0.1$ \\
& 25&  $56.8\pm1.9$ & $52.1\pm1.6$& $76.8\pm0.6$&  $\bm{81.5\pm0.4}^\dagger  $& $\bm{90.3\pm0.3}$& $\bm{90.4\pm0.3}$ \\

\midrule

\multirow{4}{*}{Proposed} 
& 1&  $\bm{57.3\pm0.6}^\dagger$ & $\bm{52.6\pm0.3}^\dagger$& $\bm{76.7\pm0.4}^\dagger$& $\bm{80.0\pm0.3}^\dagger$& $\bm{87.9\pm0.1}^\dagger$& $\bm{88.7\pm0.1}^\dagger$  \\
& 5&  $\bm{62.9\pm1.0}$ & $\bm{61.0\pm0.9}^\dagger$& $\bm{80.7\pm0.6}^\dagger$&  $\bm{80.7\pm0.0}$& $89.5\pm0.1$& $90.3\pm0.0^\dagger$ \\
& 10&  $\bm{64.3\pm1.5}^\dagger$ & $\bm{63.0\pm0.3}^\dagger$& $\bm{82.2\pm0.1}^\dagger$&  $\bm{81.5\pm0.1}^\dagger$& $\bm{90.6\pm0.1}^\dagger$& $\bm{90.8\pm0.1}^\dagger$ \\
& 25&  $\bm{64.5\pm1.3}$ & $\bm{63.4\pm1.2}$& $\bm{82.1\pm0.5}^\dagger$&  $81.3\pm0.2$& $90.1\pm0.2$& $90.3\pm0.1$ \\

\midrule
\multirow{1}{*}{Full data}
& 100 & $64.1\pm0.7$ & $66.4\pm1.0$ & $83.1\pm0.5$ & $81.4\pm0.2$ & $89.7\pm0.2$ & $90.2\pm0.1$
 \\

\bottomrule
\end{tabular}

}
\caption{Performance comparison across distillation ratios ($\eta$) on TUEV and TUAB datasets. $\bm\kappa$ refers to Cohen’s kappa coefficient. Bold values indicate the best performance within each distillation ratio, excluding the full data case. DS refers to the diversity sampling. The $^\dagger$ symbol indicates the result is significantly better ($p<0.05$) than the corresponding Random baseline at the same $\eta$ based on Mann–Whitney U test.}
\label{tab:distillation-results-2}
\end{table*}

The proposed framework utilizes the selection-based distillation method. We formalize the process of selecting the most representative EEG samples as the coreset construction problem.
Formally, let $\mathcal{Z}\subset \mathbb{R}^{N\times d}$ be the EEG compressed pre-training dataset, where $N$ is the sample size, $d$ is the dimension of compressed representations. 
The objective is to select $k$ points from $\mathcal{Z}$ such that every data point $\mathbf{z}$ is closest to one selected center, and the largest distance between any point to its closest center is minimized, which is described as

\begin{equation}
\label{eqn:k-center}
    \min_{\bm\mu \subset \mathcal{Z}}\max_{\mathbf{z}\in \mathcal{Z}}\min_{k\in \mathcal{K}}  ||\mathbf{z} - \bm{\mu}_k||_2^2.
\end{equation}

This problem is NP-hard, so we utilize greedy approximation solver proposed in \cite{sener2018active} to iteratively select points that maximize the minimum distance to previously chosen k-centers, of which the time complexity is $\mathcal{O}(k \times N\times d)$.

\begin{figure}[!t]
    \centering
    \includegraphics[width=0.80\linewidth]{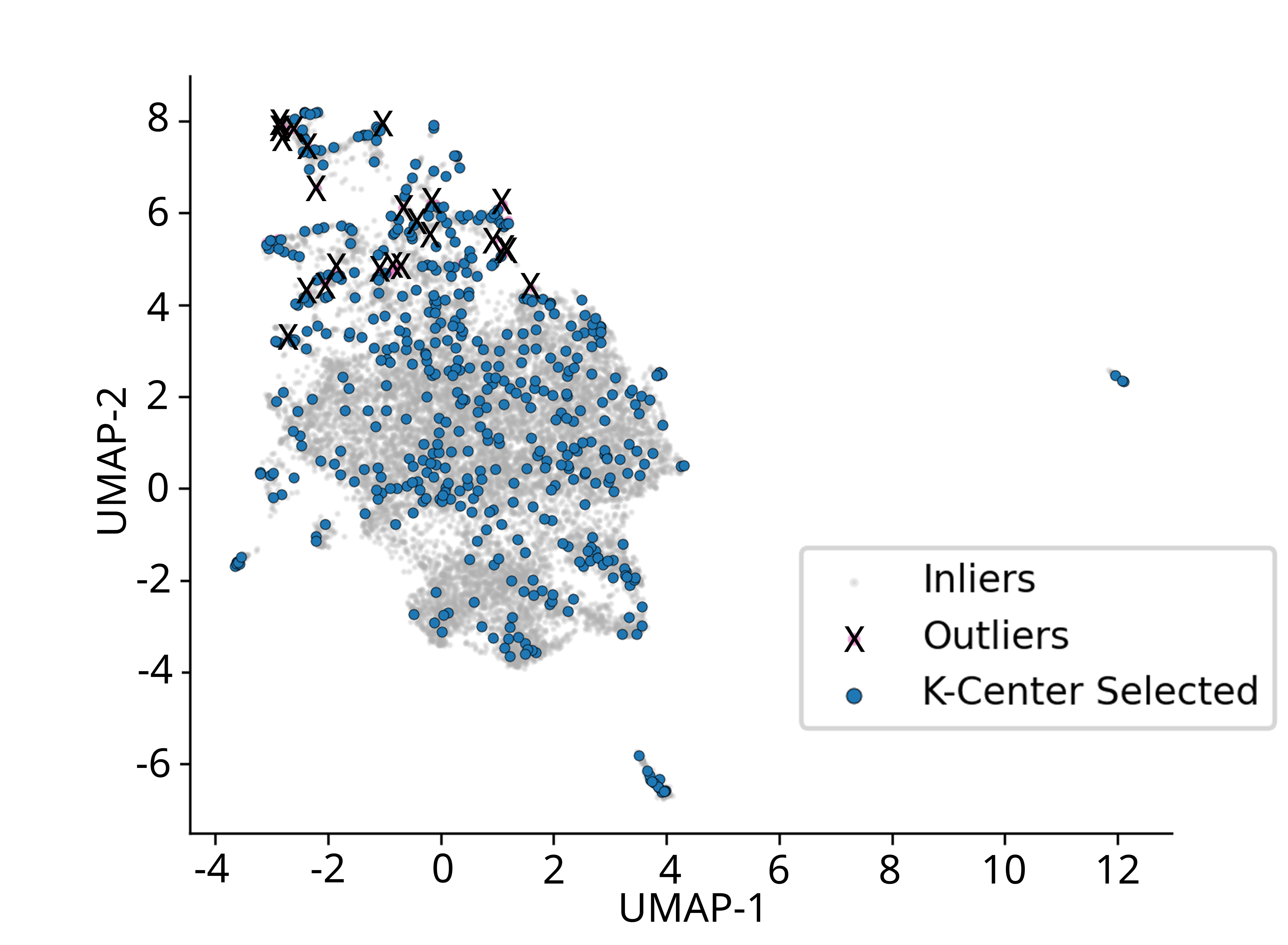}
    \caption{UMAP visualization of compressed EEG segments after OOD removal and diversity sampling on SEED-sleep~\cite{seed-sleep}.}
    \label{fig:umap}
\end{figure}

\section{Experiments \& Results}
This section elaborates on the baselines, implementation details, the setup during pre-training and fine-tuning stages, followed by the results and in-depth analysis.

\subsection{Baselines}
We compare the proposed framework with three baselines under varying distillation ratios.

\textbf{Random} baseline uniformly samples data without considering structure, serving as a simple, unbiased lower-bound baseline.

\textbf{PCA}~\cite{Anuragi2024,Fujiwara2020} reduces EEG dimensionality by projecting data onto top principal components. We use Incremental PCA to scale efficiently and assess how it compares with SSL-based embeddings in representing EEG diversity.

\textbf{M3D}~\cite{zhang2024m3ddatasetcondensationminimizing} generates synthetic data by minimizing embedding discrepancies using random, untrained networks. Although it is lightweight and computationally feasible at low distillation ratios, it still becomes impractical at 25 percent distillation ratio due to excessive GPU memory demands.

\subsection{Implementation Details}
The input signals from both the potential and spectral domains are independently normalized and segmented into 20 non-overlapping patches. Then, the neural compressor is trained for 50 epochs using the Adam optimizer, with the learning rate of 0.001 and gradient clipping applied at a maximum norm of 5.0. A scheduler is used to reduce the learning rate every 10 epochs with a decay factor of 0.5. The training objective, described in Equation \ref{eqn:training-obj}, incorporates an instance discrimination loss term $\mathcal{L}_\text{IDC}$, weighted by $\beta = 0.0001$. The encoder consists of 6 self-attention layers with 8 heads per layer, while the decoder contains 2 transformer layers, each with 8 attention-heads as well. Each EEG segment $X$ is projected into a 64-dimensional embedding space.

\subsection{Experiment Setup} 
\paragraph{Dataset Construction}
We use the same pre-training datasets as in LaBraM \cite{jiang2024largebrainmodellearning}, which includes 32 datasets over 2,500 hours of EEG data collected from multiple tasks (e.g., emotion recognition, motor imagery, etc.). These datasets have diverse channel montages. Each EEG sample is a segment into $X \in \mathbb{R}^{c \times T}$, where $c$ is the number of channels and $T$ is the temporal length (either 4 or 8 seconds), with a fixed stride of 4 seconds. All data are band-pass filtered within the range [0.1, 75] Hz and resampled to 200Hz. During the distillation process, each dataset is distilled independently. 

\begin{figure}[!b]
    \centering
    \includegraphics[width=0.75\linewidth]{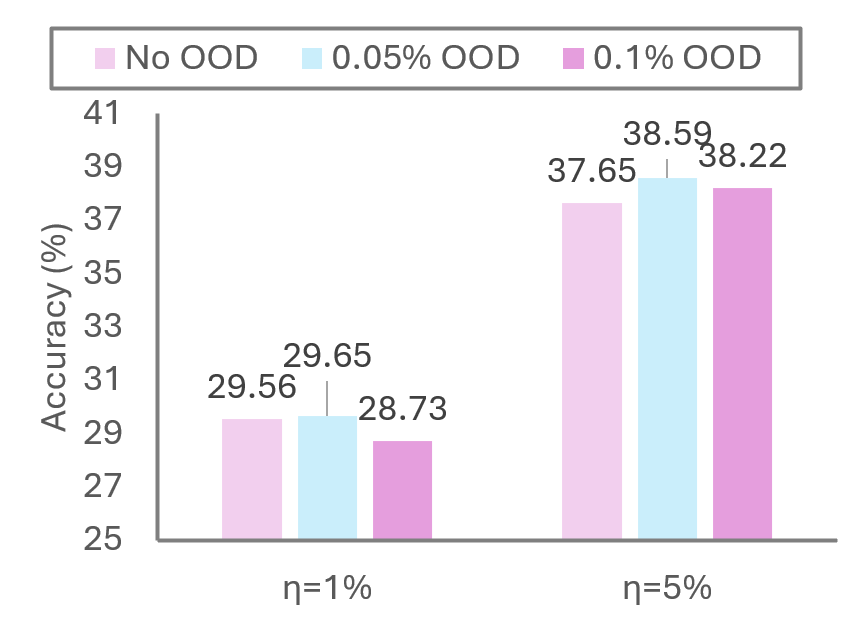}
    \caption{Impact of outlier removal on SEED-V accuracy at different distillation ratios $\eta$ and OOD removal thresholds $\tau$.}
    \label{fig:ood-removal}
\end{figure}

\paragraph{Pre-training Configuration}
We adopt the LaBraM-base architecture, consisting of 12 encoder layers with a hidden size of 200, 10 attention heads, and a multi-layer-perceptron dimension of 800. The model is trained using the AdamW optimizer with $\beta_1 = 0.9, \beta_2 = 0.98$, and a cosine learning rate scheduler with the weight decay of 0.05. The pre-training objective follows the symmetric masking formulation $\mathcal{L} = \mathcal{L}_M + \mathcal{L}_M^{\text{sys}}$. All the pre-training and fine-tuning experiments are conducted on four NVIDIA RTX 4090 GPUs.

\subsubsection{Downstream Evaluation}

To assess the effectiveness of each distillation method, we evaluate the pretrained foundation models on diverse types of downstream tasks. To perform rigorous comparison, each method distills the original pre-training dataset at multiple compression ratios ($\eta = 1\%, 5\%, 10\%, 25\%$) first, and the resulting distilled subsets are used for pre-training. Finally, the pretrained models are fine-tuned separately on four representative downstream tasks that cover both classification and regression settings. We follow the identical evaluation protocol established in LaBraM to ensure direct comparability. All experiments use consistent hyperparameters during pre-training and fine-tuning. Results are averaged across five random seeds, and standard deviations and results' significance are reported in Table \ref{tab:distillation-results} and \ref{tab:distillation-results-2}. Below are the details of downstream tasks:
\begin{itemize}
\item \textbf{TUEV}~\cite{10.3389/fnins.2016.00196}: Six-class classification of EEG events.
\item \textbf{TUAB}~\cite{10.3389/fnins.2016.00196}: Binary classification of normal versus abnormal EEG signals.
\item \textbf{SEED-V}~\cite{SEED-V}: Five-class emotion recognition based on EEG.
\item \textbf{MoBI}~\cite{mobi}: Continuous regression of bilateral lower-limb joint angles from EEG during walking.
\end{itemize}


\subsection{Distillation Performance}

\paragraph{The foundation model retains comparable performance with ~5\% of the pre-training data.} As the dataset percentage increases, the performance on all the downstream tasks demonstrates improvements, indicating enhanced model generalization with more training data, and the performance change is also aligned with the experiments reported in \cite{jiang2024largebrainmodellearning}. Tables \ref{tab:distillation-results} and \ref{tab:distillation-results-2} indicate a consistent pattern of data redundancy in large-scale EEG pre-training datasets. Specifically, we observe that using only ~5\% of the pre-training data selected by EEG-DLite achieves performance close to the upper bound. In contrast, the random selection baseline requires around 25\% of the data to reach a similar level. These results suggest that a substantial portion of the data has limited impact on model performance, highlighting the potential for more efficient training and deployment through targeted data selection.

\paragraph{EEG-DLite consistently outperforms across all downstream datasets and distillation ratios.}
Across all evaluated datasets and ratios, the proposed framework consistently outperforms the random sampling and M3D baselines, demonstrating its effectiveness in selecting informative and diverse subsets. These results highlight the importance of maintaining diversity during distillation to enhance generalization and model robustness, even with limited data. Furthermore, we find that SSL generates more stable and discriminative representations than PCA under the same output dimensionality. On datasets such as TUEV, MoBI, and TUAB, models trained on distilled data even surpass those trained on the full dataset, indicating that careful sample selection could be more effective than brute-force scaling. Although the EEG segments are unlabeled, the UMAP visualization in Figure~\ref{fig:umap} reveals distinct clustering patterns in the latent space, reflecting the inherent structure and diversity present in EEG signals as well.



\begin{table}[!h]
\centering
\small
\setlength{\tabcolsep}{4pt}
\begin{tabular}{lccccc}
\toprule
\textbf{Method} & \textbf{$\bm{\eta}$ (\%)} & \textbf{Acc.} & \textbf{F1} & $\bm\kappa$ \\
\midrule
\multirow{2}{*}{Random} 
& 50  & 53.4 & 54.0 & 28.9 \\
& 25  & 52.8 & 52.1 & 28.0 \\
\midrule
\multirow{2}{*}{PCA + DS} 
& 50  & 51.8 & 48.6 & 27.4 \\
& 25  & 51.6 & 49.9 & 26.2 \\
\midrule
\multirow{3}{*}{Proposed ($\tau=0$)} 
& 50  & 54.1 & 52.8 & 29.9 \\
& 25  & 54.6 & 55.1 & 29.1 \\
\midrule
\multirow{2}{*}{Proposed ($\tau=1\%$)} 
& 50  & $\textbf{56.6}$ & $\textbf{56.7}$ & $\textbf{33.2}$ \\
& 25  & $\textbf{55.3}$ & $\textbf{55.7}$ & $\textbf{31.3}$ \\
\midrule
{Full Data} & 100 & 54.6 & 55.4 & 30.8 \\
\bottomrule
\end{tabular}
\caption{Pilot study evaluating EEG-DLite on SEED using EEGNet in the cross-subject supervised setting at different distillation ratios $\eta$ and OOD removal ratios $\tau$.}
\label{tab:pilot-study}
\end{table}

\paragraph{OOD removal is helpful.}
Figure~\ref{fig:ood-removal} presents an ablation study assessing the effect of varying the outlier removal threshold $\tau$. The results show that removing the top 0.05\% samples with the highest OOD scores leads to improved downstream accuracy. Manual inspection reveals that these samples often contain noise or signal artifacts, which can degrade representation quality.

To further evaluate the effectiveness of EEG-DLite, we conduct a pilot study on SEED dataset~\cite{zheng2015investigating, duan2013differential} using a small network like EEGNet in the cross-subject supervised learning setting. The results in Table~\ref{tab:pilot-study} show that EEG-DLite achieves the highest performance when trained on only 25\% of the data, even outperforming models trained on the full dataset. We also observe the model is sensitive to the presence of OOD samples in the supervised setting.

\begin{figure}[!t]
    \centering
    \includegraphics[width=0.75\linewidth]{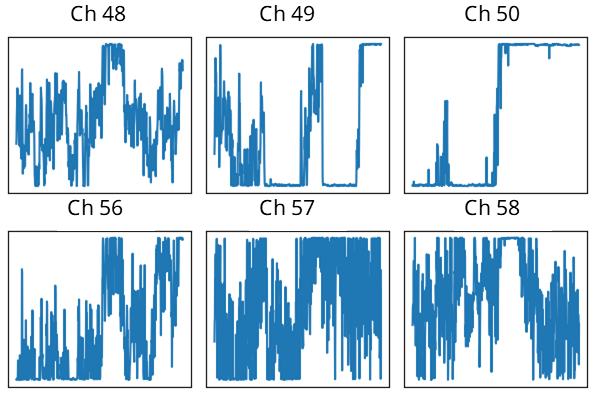}
    \caption{Example of generative EEG sample with M3D.}
    \label{fig:m3d-segment}
\end{figure}

\paragraph{Large-scale EEG datasets pose new challenges to the data synthesis approach.}
Although generative approaches have shown effectiveness in computer vision tasks~\cite{zhang2024m3ddatasetcondensationminimizing, zhao2022datasetcondensationdistributionmatching}, their effectiveness on physiological signals such as EEG remains largely unexplored. In our experiments, M3D, a light-weighted synthesizing framework, was selected considering the huge computational effort to generate large pre-training datasets. M3D consistently underperforms compared to all other baselines, including random sampling. As illustrated in Figure~\ref{fig:m3d-segment}, the generated EEG segments exhibit unnatural characteristics, such as abrupt plateaus, flat transitions, and repetitive blocky patterns, all of which deviate from realistic brain signal dynamics. These artifacts reflect poor temporal and spectral fidelity. Furthermore, the generative process in M3D is computationally intensive, which poses scalability challenges for large-scale applications. Taken together, these findings highlight that for EEG and related physiological signals, both the synthetic quality and computational feasibility are important considerations. 

\begin{figure}[!t]  
    \centering
    \begin{minipage}{\linewidth}
        \centering
        \includegraphics[width=0.65\linewidth]{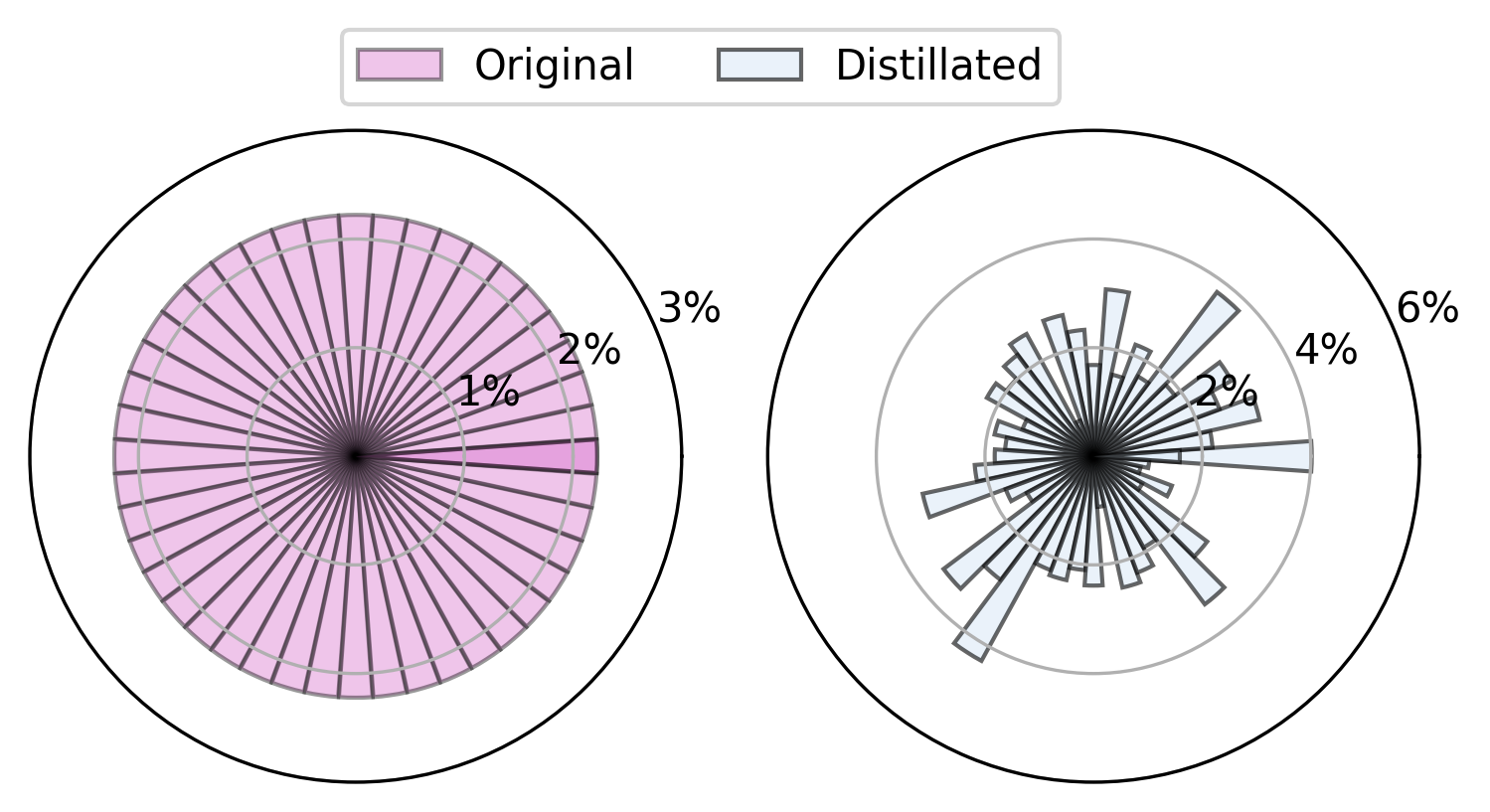}
        \label{fig:seed3}
    \end{minipage}
    \begin{minipage}{\linewidth}
        \centering
        \includegraphics[width=0.65\linewidth]{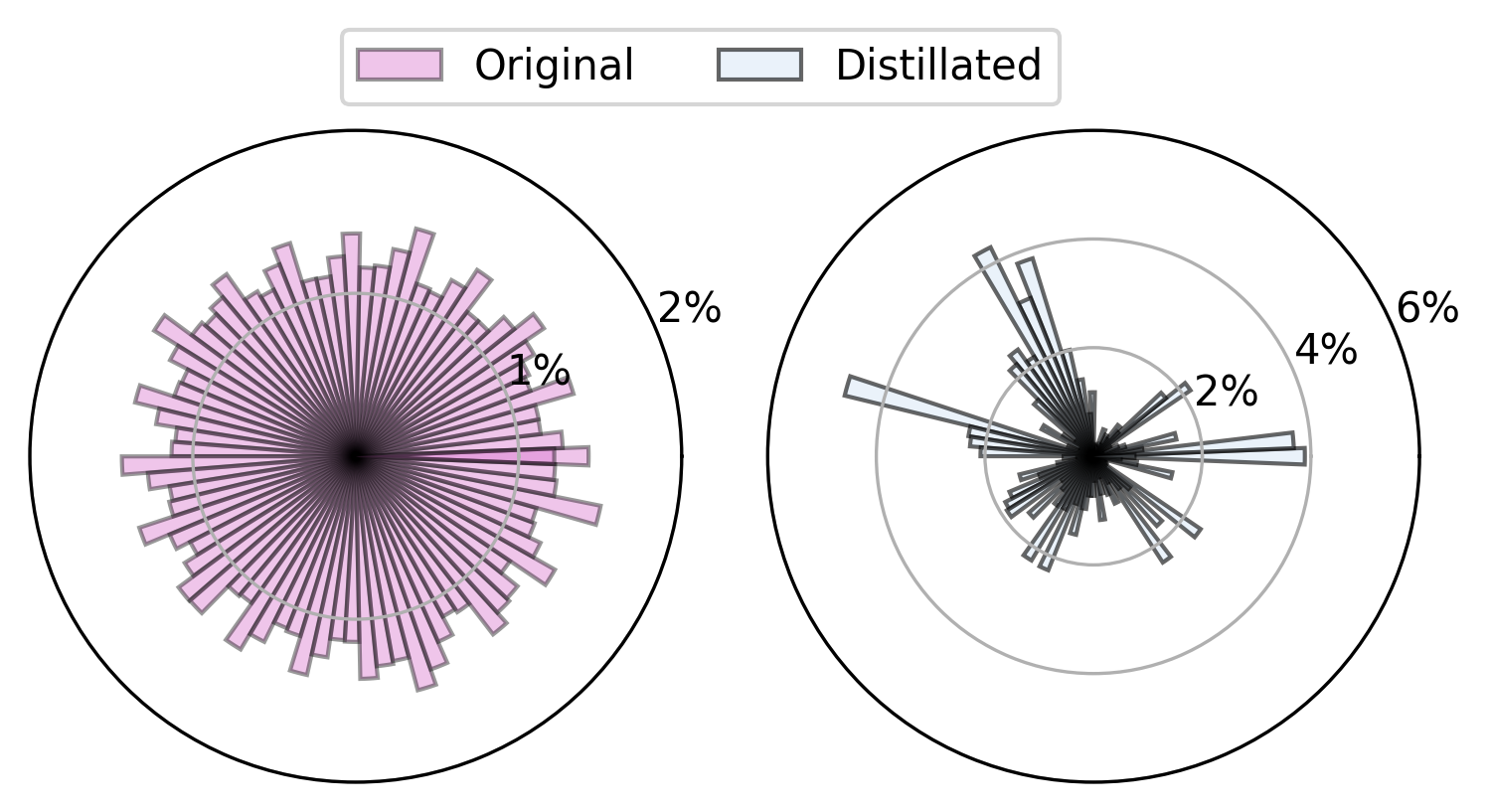}
        \label{fig:seed7}
    \end{minipage}
    \caption{Sample distribution per subject in SEED~\cite{seed} (top) and SEED-VII~\cite{seed-7} (bottom). Each bin represents a subject, and its height indicates the proportion of samples contributed by that subject in the original or 5\% distilled dataset.}
    \label{fig:subject-sample-dist}
\end{figure}

\paragraph{Subject variance becomes significant after diversity sampling.}
BCI datasets typically involve recordings from multiple subjects, each contributing EEG segments with varying quality and characteristics. After applying diversity sampling, the percentage of samples per subject shows notable variation. As illustrated in Figure~\ref{fig:subject-sample-dist}, some subjects contribute significantly more segments than others. This imbalance reflects inherent inter-subject variability in EEG signals, which can stem from differences in neural dynamics, recording conditions, or noise levels. Recognizing and understanding these subject-level patterns may inform several directions for future research. For instance, it may enable subject-aware pre-training strategies. 

\section{Conclusion}
In this study, we propose a novel data distillation framework to condense large-scale unlabeled EEG datasets through three key steps: compression, outlier removal, and diversity sampling. The framework is evaluated across four downstream tasks at varying distillation ratios, demonstrating that only 5\% of the original data is sufficient to train a foundation model with comparable performance. These results highlight that data quality and diversity contribute more significantly to the generalization ability of foundation models than data quantity. We believe EEG-DLite represents a practical step toward efficient foundation model pre-training and provides insights about diverse dataset design and selection for broader downstream applications.

\section{Acknowledgments}
This work is supported by the MOE Tier 2 Project (MOE-T2EP20124-0001).

\bibliography{aaai2026}

\end{document}